\documentclass[conference]{IEEEtran}
\IEEEoverridecommandlockouts

\usepackage{tikz}

\usepackage{amsmath}
\usepackage[normalem]{ulem}
\usepackage{amssymb}
\usepackage{amsfonts}
\usepackage{multicol}
\usepackage{graphicx}
\usepackage{pifont}
\usepackage[noend]{algorithmic}
\usepackage[ruled, linesnumbered]{algorithm2e}
\usepackage{enumitem}
\usepackage{multirow}
\usepackage{color}
\usepackage{listings}
\usepackage{float}
\usepackage{ragged2e}
\usepackage{xspace}
\usepackage{soul}
\usepackage{lipsum}
\usepackage{tabularx}
\usepackage{makecell}
\usepackage{mathtools}
\usepackage{subcaption}
\usepackage{fancyhdr}
\usepackage[normalem]{ulem}
\usepackage{filecontents}
\usepackage{tikz}
\usetikzlibrary{decorations.pathmorphing}
\usetikzlibrary{arrows}
\usetikzlibrary{positioning}
\usepackage{standalone}
\usepackage{booktabs}

\usepackage{cite}
\usepackage{textcomp}
\usepackage{xcolor}

\newcommand{\Fig}[1]{Fig.~\ref{#1}}

\newcommand{\Tbl}[1]{Tbl.~\ref{#1}}
\newcommand{\Sec}[1]{Sec.~\ref{#1}}

\newcommand{\benchmark}[1]{{\texttt{#1}}}
\renewcommand{\paragraph}[1]{\vspace*{0.15cm}\noindent\textbf{#1}\hspace*{.1cm}}

\newcommand*\circled[1]{\tikz[baseline=(char.base)]{
               \node[shape=circle,fill,inner sep=0.6pt] (char) {\textcolor{white}{#1}};}}

\definecolor{codegreen}{rgb}{0,0.6,0}
\definecolor{codegray}{rgb}{0.5,0.5,0.5}
\definecolor{codepurple}{rgb}{0.58,0,0.82}
\definecolor{backcolour}{rgb}{0.95,0.95,0.92}

\lstdefinestyle{mystyle}{
    backgroundcolor=\color{backcolour},   
    commentstyle=\color{codegreen},
    keywordstyle=\color{magenta},
    numberstyle=\tiny\color{codegray},
    stringstyle=\color{codepurple},
    basicstyle=\ttfamily\footnotesize,
    breakatwhitespace=false,         
    breaklines=true,                 
    captionpos=b,                    
    keepspaces=true,                 
    numbers=left,                    
    numbersep=5pt,                  
    showspaces=false,                
    showstringspaces=false,
    showtabs=false,                  
    tabsize=2
}

\def\BibTeX{{\rm B\kern-.05em{\sc i\kern-.025em b}\kern-.08em
    T\kern-.1667em\lower.7ex\hbox{E}\kern-.125emX}}
\begin{document}

\title{Nesting Forward Automatic Differentiation for Memory-Efficient Deep Neural Network Training
\thanks{$^*$ Jingwen Leng and Minyi Guo are corresponding authors of this paper.}
}

\author{\IEEEauthorblockN{Cong Guo$^{1,2}$, Yuxian Qiu$^{1,2}$, Jingwen Leng$^{1,2,*}$, Chen Zhang$^{3}$  \\Ying Cao$^{4}$, Quanlu Zhang$^{4}$, Yunxin Liu$^{5}$, Fan Yang$^{4}$, Minyi Guo$^{1,2,*}$}
\IEEEauthorblockA{\textit{$^{1}$Shanghai Jiao Tong University, $^{2}$Shanghai Qi Zhi Institute, $^{3}$Alibaba Group}}
\IEEEauthorblockA{\textit{$^{4}$Microsoft Research, $^{5}$Institute for AI Industry Research (AIR), Tsinghua University}}
}

\maketitle

\begin{abstract}
    An activation function is an element-wise mathematical function and plays a crucial role in deep neural networks (DNN). 
   Many novel and sophisticated activation functions have been proposed to improve the DNN accuracy but also consume massive memory in the training process with back-propagation.
    In this study, we propose the nested forward automatic differentiation (Forward-AD), specifically for the element-wise activation function for memory-efficient DNN training.
    We deploy nested Forward-AD in two widely-used deep learning frameworks, TensorFlow and PyTorch, which support the static and dynamic computation graph, respectively.
    Our evaluation shows that nested Forward-AD reduces the memory footprint by up to 1.97$\times$ than the baseline model and outperforms the recomputation by 20\% under the same memory reduction ratio.
\end{abstract}

\section{Introduction}
\label{sec:introduction}

Deep neural network (DNN) models have gained tremendous success in many important domains. 
For example, ResNet~\cite{he2016deep} and BERT~\cite{devlin2018bert} (based on pre-trained Transformer network~\cite{vaswani2017attention}) have shown impressive accuracy in the challenging area of image classification~\cite{deng2009imagenet} and natural language processing~\cite{manning1999foundations} (NLP) tasks.
Researchers have shown that activation functions are important elements in DNN models, and proposed many novel activation functions for the better accuracy.
For most vision tasks, Mish~\cite{misra2019mish}, Swish~\cite{ramachandran2017searching} and GELU~\cite{hendrycks2020gaussian} surpass ReLU, specifically, about 1\% accuracy improvement on ResNet~\cite{misra2019mish, ramachandran2017searching}. 
On the other hand, GELU~\cite{hendrycks2020gaussian} is the most widely used activation function in the NLP models and achieves the best accuracy among other candidates. 

\begin{figure}[t]
  \centering
  \includegraphics[width=1\columnwidth]{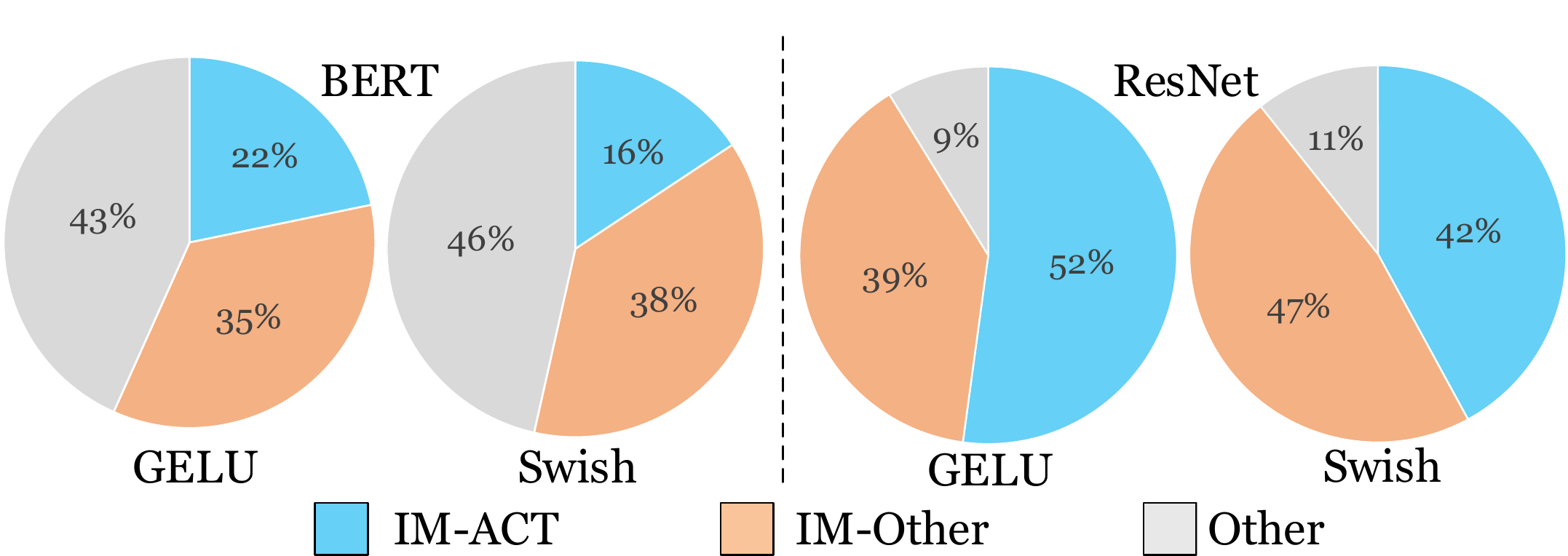}
  \caption{Memory breakdown for DNN model training on PyTorch. \benchmark{IM-ACT} is  the intermediate variable footprint of activation function, and \benchmark{IM-Other} is the rest intermediate variable footprint. \benchmark{Other} includes the weights and workspaces for the temporary variables, such as gradients. The batch size of all these experiments is set to 16. 
  }
\label{fig:act_memory}
\end{figure}

The activation functions mentioned above are characterized by their sophisticated architectures and massive memory consumption, especially when combined with straightforward implementations. In some cases, they become the top memory consumers, surpassing the intermediate variables of convolution or fully connected layers saved in the forward pass for gradient computation.
The employment of novel activation functions exacerbates the model applicability when training with modern commodity accelerators, such as GPUs, with limited global memory capacity. For example, the activation-related variables of BERT-base and ResNet-50 occupy 22\% and 52\% memory footprint, respectively, as shown in \Fig{fig:act_memory}. 

Many memory optimization approaches~\cite{jain2019checkmate, rhu2016vdnn, rhu2018compressing, chen2016training} have been proposed, but none of them is appropriate for activation functions without introducing extra computing overhead. 
To reduce activation functions' footprint usage during model training, recomputation~\cite{chen2016training, jain2019checkmate} reproduces the intermediate variables in the backward pass without saving them in memory. 
However, the recomputation inserts new operators breaking the original computation graph and brings extra computing overhead, dropping the speed of training. 

We introduce forward mode automatic differentiation (Forward-AD, FAD) for gradient computation of activation functions to avoid the recomputation overhead and reduce the memory footprint. 
Automatic differentiation~\cite{GunesBaydin2018} (autodiff, AD) is a family of mathematics tools to automatically and accurately evaluate numeric function derivatives using computer programs. 
There are two modes of autodiff: backward mode (i.e., back-propagation, BP) and forward mode (FAD). 

Back-propagation is the mainstay approach for DNN training.
Today's mainstream machine learning frameworks, such as PyTorch~\cite{paszke2017automatic} and TensorFlow~\cite{abadi2016tensorflow}, implement BP using dynamic/static computational graphs and significantly improve model deployment efficiency. 
Compared to back-propagation AD, forward AD can reduce the stored intermediate variables and execute efficiently and straightforward in the specific numeric function $f$: $\mathbb{R}^N \rightarrow \mathbb{R}^M,\  (N \leq M)$. 
The activation functions are the typical element-wise functions with $f$: $\mathbb{R}^1 \rightarrow \mathbb{R}^1$, whose computation graphs can be optimized by FAD to reduce the DNN training's memory consumption.

We propose an element-wise specific computation graph optimization, which substitutes the original BP execution of sub-graphs with nested FAD by recognizing the original graph's specific function pattern ($\mathbb{R}^1 \rightarrow \mathbb{R}^M$) without influencing on the remaining graph. 
This approach can automatically optimize both static and dynamic computation graphs in the popular deep learning frameworks, especially in the imperative mode in PyTorch, and achieves memory reduction for DNN training.

The contribution of our work is as follows:

\begin{itemize}
\item We propose a memory optimization approach with the nested FAD inside BP in the deep learning framework.
\item  The nested FAD can automatically be executed in the popular frameworks with the two execution modes: dynamic and static computation graphs.
\item  We evaluate FAD in the state-of-the-art models achieving a higher memory reduction ratio (as high as 1.97$\times$ and 1.34$\times$ on average) than the original end-to-end model and an average of 1.78$\times$ speedup than recomputation for the activation function on BERT.
\end{itemize}

We organize the paper as follows. \Sec{sec:forwardad} introduces the background of the automatic differentiation and the overview of the nested FAD algorithm. 
We explain nested FAD implementation with static and dynamic computation graph in \Sec{sec:optimization} and \Sec{sec:dynamic_graph}, respectively.
We evaluate the nested FAD with DNN models in \Sec{sec:evaluation}, introduce the related work in \Sec{sec:related_work}, and conclude in \Sec{sec:conclude}.

\section{Activation and Autodiff}
\label{sec:forwardad}

This section first introduces the relevant background on the activation function and analyzes the memory efficiency of the activation function in DNN models. Then, we compare the differences among BP, recomputation, and FAD with a specific example. Finally, we define the nested FAD in the BP algorithm of the DNN models.

{
\renewcommand{\arraystretch}{1.5}
\begin{table}[b]
\centering  
\resizebox{1\linewidth}{!}{
\begin{tabular}{c | c } 
\Xhline{1pt} 
\textbf{GELU} & $0.5\cdot x\cdot \{1+\tanh[\sqrt{2/\pi}\cdot (x + 0.044715 \cdot x^3)]\}$\\\hline
\textbf{Mish} & $x\cdot \tanh[\ln(1+e^x)] = x\cdot \tanh[\text{softplus}(x)]$    \\\hline
\textbf{Swish} & $x/(1+e^{-x}) =  x\cdot \text{sigmoid}(x)$ \\
\Xhline{1pt}
\end{tabular}
}
\caption{Activation functions.}
\label{tbl:activation}
\end{table}
}

\subsection{Activation functions}
\label{background:activation}
DNN models have recently achieved state-of-the-art results in many important domains,  e.g., convolution neural network~\cite{lecun1995convolutional} (CNN) in the computer vision domain, and BERT~\cite{devlin2018bert} in the natural language processing domain.
Recently, many activation functions, e.g., Mish~\cite{misra2019mish}, Swish~\cite{ramachandran2017searching}, and GELU~\cite{hendrycks2020gaussian}, have been proposed to optimize ResNet-50~\cite{he2016deep} and BERT~\cite{vaswani2017attention}, showing higher accuracy than ReLU. Their formulas are shown in \Tbl{tbl:activation}.

\subsection{Footprint of activation}
\label{background:Footprint}
BP algorithm needs to save intermediate variables in the forward pass for computing gradients in the backward pass. 
The amount of weight is fixed, and the gradients can be released immediately after computation. Only the saved intermediate variables would persist in the memory for a long time and increase with the batch size. 

We collect the memory footprint usage in ResNet-50 and BERT with activation functions GELU and Swish in PyTorch. 
The two activations have different intermediate variables usage due to their different derivative functions.
The memory breakdown results in \Fig{fig:act_memory} show great potential for memory optimization utilizing FAD, which, without the burden of saving intermediate variables, has much less memory footprint than back-propagation.  It is noteworthy that the proportion of \benchmark{IM-ACT} (and \benchmark{IM-Other}) will increase as the batch size increase. That will strengthen the performance of Forward-AD.

\subsection{Forward-AD, BP and recomputation}
\begin{figure*}[t]
  \centering
  \includegraphics[width=2\columnwidth]{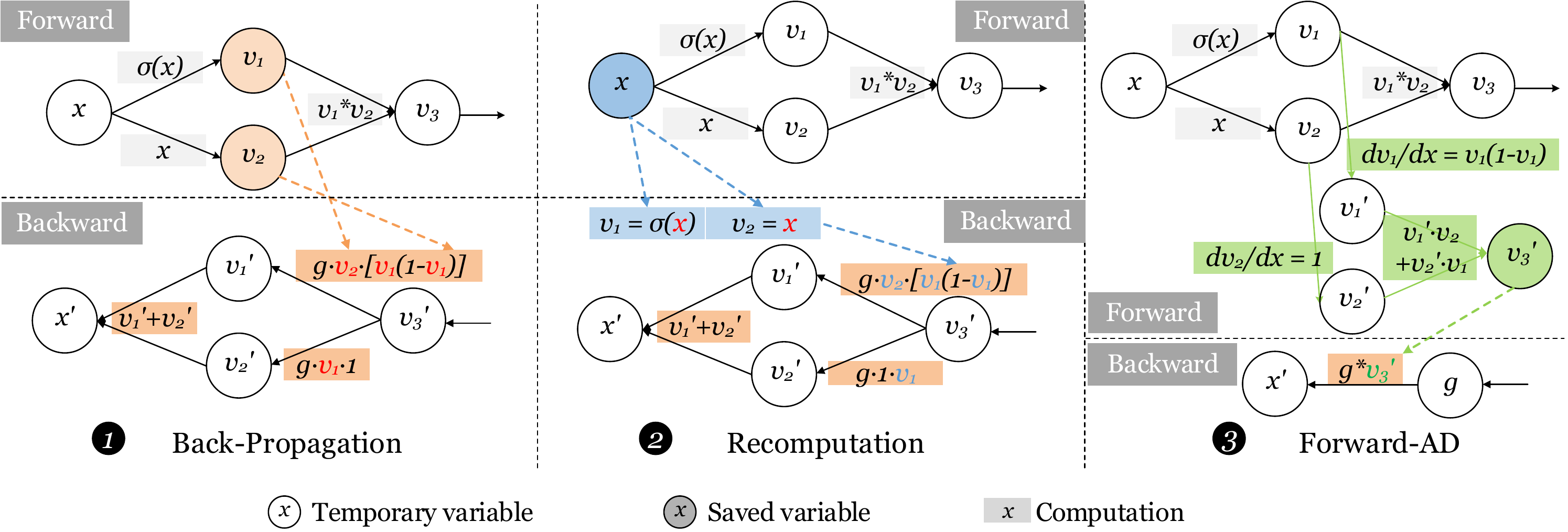}
  \vspace*{0.04cm}
  \caption{Comparison of forward and backward computation graphs for back-propagation AD, recomputation, and forward AD.}
\label{fig:fad}
\end{figure*}

We explain the difference between recomputation and FAD using the Swish activation function as an example. The formula of Swish is shown in \Tbl{tbl:Swish}. Swish function is a simple activation that has two operations: \benchmark{Mul} and \benchmark{Sigmoid}.
\Fig{fig:fad} depicts the three approaches: BP, recomputation, and FAD with an input value $x$.

{
\renewcommand{\arraystretch}{1.5}
\begin{table}[b]
\centering  
\resizebox{1\linewidth}{!}{
\begin{tabular}{c | c | c} 
\Xhline{1pt} 
 & Forward pass & Backward pass (derivative)\\\hline
\textbf{Swish} & $x\cdot \sigma(x)$ & $\sigma(x) + x \cdot \sigma(x) \cdot [1 - \sigma(x)]$ \\
\Xhline{1pt}
\end{tabular}
}
\caption{Swish function. $\sigma(x) = Sigmoid(x)$.}
\label{tbl:Swish}
\end{table}
}

\paragraph{Original BP (\circled{1}) algorithm} first executes the forward pass with three nodes: 
$$v_1 = \sigma(x), \ v_2 = x, \ v_3 = v_1\cdot v_2.$$

In backward pass, Node $v_3'$ receives its gradient $g$ from upstream and calculates gradient for $v_1'$ and $v_2'$. 
Obviously, back-propagation needs to save intermediate variables $v_1$ and $v_2$ for computing Swish gradient. Node $v_3'$ will deliver downstream gradients $g\cdot v_1$ to Node $v_2'$, and $g\cdot v_2$ to Node $v_1'$. Then, BP continues to traverse $v_1'$ and $v_2'$ recursively.
Node $v_1'$ calculates the gradient by
\begin{equation}
  \begin{aligned}
  v_1' =& \ g \cdot v_2 \cdot [\sigma(x)(1-\sigma(x))]
   = \ g\cdot v_2 \cdot [v_1(1-v_1)],
  \end{aligned}
\end{equation}
based on the $\sigma(x)$ derivative, $\sigma(x)' = \sigma(x)\cdot(1-\sigma(x)).$
Node $v_2'$ calculates the gradient $v_2' = g \cdot v_1 \cdot 1$ for $x'$. Finally, Node $x'$ has the gradient
\begin{equation}
  \begin{aligned}
  x' =& v_1' + v_2'  = g\cdot v_2 \cdot [v_1(1-v_1)] + g\cdot v_1
\end{aligned}
\end{equation}
from the two branches $v_1'$ and $v_2'$.

\paragraph{Recomputation (\circled{2})} has the same forward pass as BP but releases $v_1$ and $v_2$ to reduce the memory footprint, only save the source variable $x$. Therefore, Recomputation has to recalculate the values of $v_1$ and $v_2$ in the backward pass with extra computation overhead, as show in \Fig{fig:fad} middle.

\paragraph{Forward-AD  (\circled{3})} derives and accumulates the derivative ${v_3}'$ of Swish in the forward pass and multiplies with $g$ in the backward pass. FAD simultaneously calculates Node $v_1' = v_1(1-v_1)$ and $v_2' = 1$ when Node $v_1$ and $v_2$ are executed. Finally, FAD continues to calculate 
\begin{equation}
  \begin{aligned}
  v_3' = &\ v_1\cdot v_2'  +  v_1'\cdot v_2
  = \ v_1 \cdot 1 + v_1(1-v_1) \cdot  v_2,
  \end{aligned}
\end{equation}
when the forward graph merges the Node $v_1$ and $v_2$ into $v_3$ using multiplication operation.
FAD can only store the intermediate variable $v_3'$ for back-propagation with upstream gradient $g$ and achieve the final gradient
\begin{equation}
  \begin{aligned}
    x' = &\  g \cdot v_3'
  = \ g \cdot [v_1 + v_1(1-v_1) \cdot  v_2]
  \end{aligned}
\end{equation}
directly for node $x'$.

Obviously, FAD is much more efficient than recomputation because of the memory locality without any extra computation. FAD can reduce all the intermediate variables only save the derivative $v_3'$ for the activation function instead of the variables $x$, $v_1$, and $v_2$ comparing to BP algorithm.

\subsection{Forward AD Applicability}
Autodiff can generalize the derivative of a function $f$: $\mathbb{R}^N \rightarrow \mathbb{R}^M$ by computing the Jacobian matrix $\mathbf{J}_f$ with the shape of ($M\times N$), where $M$ is the length of output vector and $N$ is for the input vector. We can compute the $\mathbf{J}_f$:
\begin{equation*}
\mathbf{J}_f = \begin{bmatrix}
                    \frac{\partial y_1}{\partial x_1} & \cdots & \frac{\partial y_1}{\partial x_N} \\
                    \vdots & \ddots & \vdots \\
                    \frac{\partial y_M}{\partial x_1} & \cdots & \frac{\partial y_M}{\partial x_N}
                  \end{bmatrix}
\end{equation*}
Forward AD computes the Jacobian–vector products (JVP): $\mathbf{J}_f\,\mathbf{v}$ and back-propagation computes the vector-Jacobian products (VJP):  ${(\mathbf{v}^T}\mathbf{J}_f)^T = \mathbf{J}_f^T\mathbf{v}$.
For example of Swish case, FAD compute the $v_1'$ and $v_2'$ with
\begin{align*}
  \mathbf{v}_{v'_{1,2}} &= [v_1',\  v_2']^T = \mathbf{J}_{f_x} \mathbf{v}_{x'} 
   =\begin{bmatrix}
    \frac{\partial v_1}{\partial x}\\
    \   \\
    \frac{\partial v_2}{\partial x} 
  \end{bmatrix} \cdot [1] ,
\end{align*}
where $x' = 1$ and compute the $v_3'$ with
\begin{align*}
  \mathbf{v}_{v'_{3}} &= [v_3'] = \mathbf{J}_{f_v} \mathbf{v}_{v'_{1,2}} 
  = \begin{bmatrix}
    \frac{\partial v_3}{\partial v_1} \  \frac{\partial v_3}{\partial v_2}
  \end{bmatrix} \begin{bmatrix}
    \frac{\partial v_1}{\partial x}\\
      \\
    \frac{\partial v_2}{\partial x} 
  \end{bmatrix} [1] \\
  &= \begin{bmatrix}
    \frac{\partial v_3}{\partial v_1}\frac{\partial v_1}{\partial x} 
    + \frac{\partial v_3}{\partial v_2}\frac{\partial v_2}{\partial x}
                    \end{bmatrix} .
  \end{align*}
For BP, we have $\mathbf{v'}_{x'} = \mathbf{J}_{f_x}^T \mathbf{v'}_{v'_{1,2}} = \mathbf{J}_{f_x}^T\mathbf{J}_{f_v}^T\mathbf{v'}_{v'_{3}}$, where $\mathbf{v'}_{v'_{3}}= [1]$.

\paragraph{Computation efficiency}
\label{s3:computation}
Evidently, for cases $f$ : $\mathbb{R}^N \rightarrow \mathbb{R}^M$, $N \leq M$, FAD is efficient for computing derivatives, and vise-versa for BP with $N > M$~\cite{wang2019demystifying}.
According to our observation, manual differentiation is widely adopted by machine learning frameworks with tremendous optimization for high performance. TensorFlow and PyTorch utilize manual differentiation to compute the derivatives and fuse operations manually. For example, layer/batch normalization, convolution, and matrix multiplication are accelerated by the highly optimized library. That optimization exploits the Jacobian matrix sparsity due to their inner \benchmark{reduction} (accumulation) operations and is incompatible with automatic differentiation, neither backward (BP) nor forward (FAD).
From the automatic differentiation perspective, the optimization is to minimize numbers of multiplication for each subregion. It is known as the Optimal Jacobian Accumulation (OJA) problem, which has been proved to be an NP-complete problem~\cite{naumann2008optimal}.

To practically implement FAD in the DNN framework, we simplify the applicability of FAD with the function $f$: $\mathbb{R}^1 \rightarrow \mathbb{R}^M$, specifically, element-wise operations with one input variable and $M \geq 1$ output variables, including most of the activation functions. 

\subsection{Nesting FAD}
We can nest FAD inside the BP algorithm for DNN models. Without loss of generality, let $v_1, v_2, ..., v_k$ be $k$ nodes in the topological ordering for the DNN computation graph $\mathbb{G}$ and ${v_k}$ is the loss $\mathbf{L}$. The gradient of the node $v_i$ computed by back-propagation is: 
\begin{equation*}
  \frac{\partial \mathbf{L}}{\partial v_i} = {\mathbf{J}_{v_i}}^T \frac{\partial \mathbf{L}}{\partial v_{i+1}}
\end{equation*}
Here, the $\frac{\partial \mathbf{L}}{\partial v_{i+1}}$ is the upstream gradient from the node $v_{i+1}$. The ${\mathbf{J}_{v_i}}$ is the Jacobian matrix of $v_i$. Assume that $v_n, ..., v_m$ in topological ordering are FAD primitive operations and they can compose to a $\mathbb{R}^1 \rightarrow \mathbb{R}^M$ element-wise operation. Then we have

\begin{equation*}
  \begin{aligned}
  \frac{\partial \mathbf{L}}{\partial v_n} &= 
  \{\prod_{i \in \{n, ..., m\}} {\mathbf{J}_{v_i}}\}^T \frac{\partial \mathbf{L}}{\partial v_{m+1}}
   = \frac{\partial {v_{m+1}}}{\partial v_n} \cdot g,
    \end{aligned}
\end{equation*}
where $g$ is the upstream gradient.
Therefore, we can exploit FAD to compute the Jacobian matrix for the element-wise operation composed by $v_n, ..., v_m$ within the forward pass. Then, FAD updates the gradients with upstream gradient $g = \frac{\partial \mathbf{L}}{\partial v_{m+1}}$.
\section{Definitions and static graph optimizaton}
\label{sec:optimization}
We optimize and nest element-wise specific FAD in two types of computation graph: static and dynamic with TensorFlow and Pytorch respectively.

\subsection{Definitions}
\label{sec:definitions}

\begin{figure}[t]
  \centering
  \includegraphics[width=0.9\columnwidth]{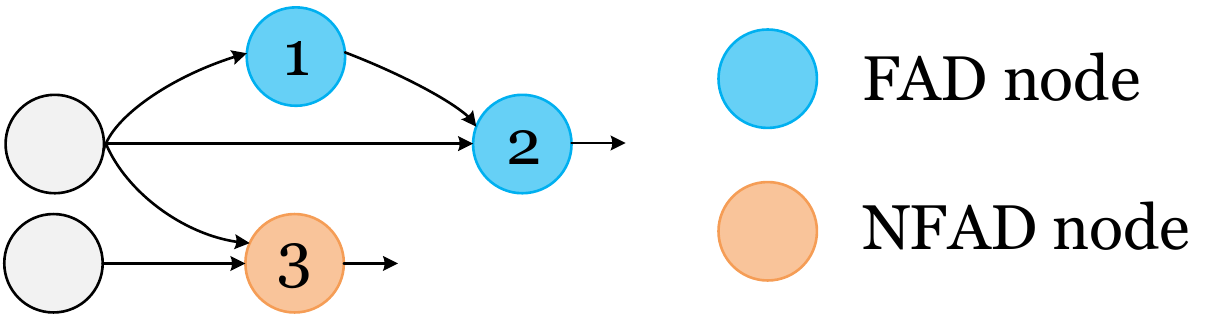}
  \vspace*{0.04cm}
  \caption{\benchmark{Fad} and \benchmark{nfad} operations.}
  \label{fig:primitive_op}
\end{figure}

We divide the operations (operator, OP) into two classes: \benchmark{fad} (FAD-compatible) and \benchmark{nfad} (FAD-incompatible). 
The primitive operations for FAD basically are \benchmark{fad} element-wise functions including:
\begin{itemize}
    \item The \benchmark{unary} operation ($f$: $\mathbb{R}^1 \rightarrow \mathbb{R}^1$) can be the primitive operation, e.g., Node 1 in \Fig{fig:primitive_op}. The \benchmark{binary} operation with one constant input variable can also be regarded as the \benchmark{unary} operation.
    \item For \benchmark{binary} operation ($f$: $\mathbb{R}^2 \rightarrow \mathbb{R}^1$), such as \benchmark{addition} and \benchmark{multiplication}, their two input elements should be originated from the same source element converting $\mathbb{R}^2 \rightarrow \mathbb{R}^1$ to $\mathbb{R}^1 \rightarrow \mathbb{R}^1$, e.g., Node 2 in \Fig{fig:primitive_op}. 
    \item Except the \benchmark{fad} operations, others operations are \benchmark{nfad}. Especially, the \benchmark{binary} operation with two tensors from different sources is the \benchmark{nfad} (FAD-incompatible) OP and violates the $f$: $\mathbb{R}^1 \rightarrow \mathbb{R}^1$, e.g., Node 3 in \Fig{fig:primitive_op}.
\end{itemize}

Manual differentiation~\cite{GunesBaydin2018} is widely used in frameworks. It can greatly impact the efficiency of the element-wise operation. For example, \benchmark{Sigmoid} is a classic activation function: $S(x) = \frac{1}{1+e^{-x}}$ and would produce lots of intermediate variables with the original format. Frameworks implement its enclosed symbolic format as a basic operator. Without loss of practicability, we define the symbolic operators implemented in frameworks as the primitive operation, including \benchmark{tanh}, \benchmark{sigmoid} and \benchmark{softplus}. 


The operation composed of finitely many primitive operations is the FAD-compatible operation and can maintain $f$: $\mathbb{R}^1 \rightarrow \mathbb{R}^M, M \geq 1$, which is efficient for FAD and the subject matter of this paper. The others are \benchmark{nfad} OPs.

\subsection{Static computation graph}
\label{sec:static_graph}

TensorFlow is the most representative static computation graph framework. It defines the graph statically before a model execution and the graph can not be modified during the run-time. That is friendly to optimize and accelerate the computation of the DNN model without convenience and flexibility for users.
Naturally, it is easier for a static graph to implement the nested FAD than a dynamic graph.

\begin{figure}[t]
    \centering
    \includegraphics[width=1\columnwidth]{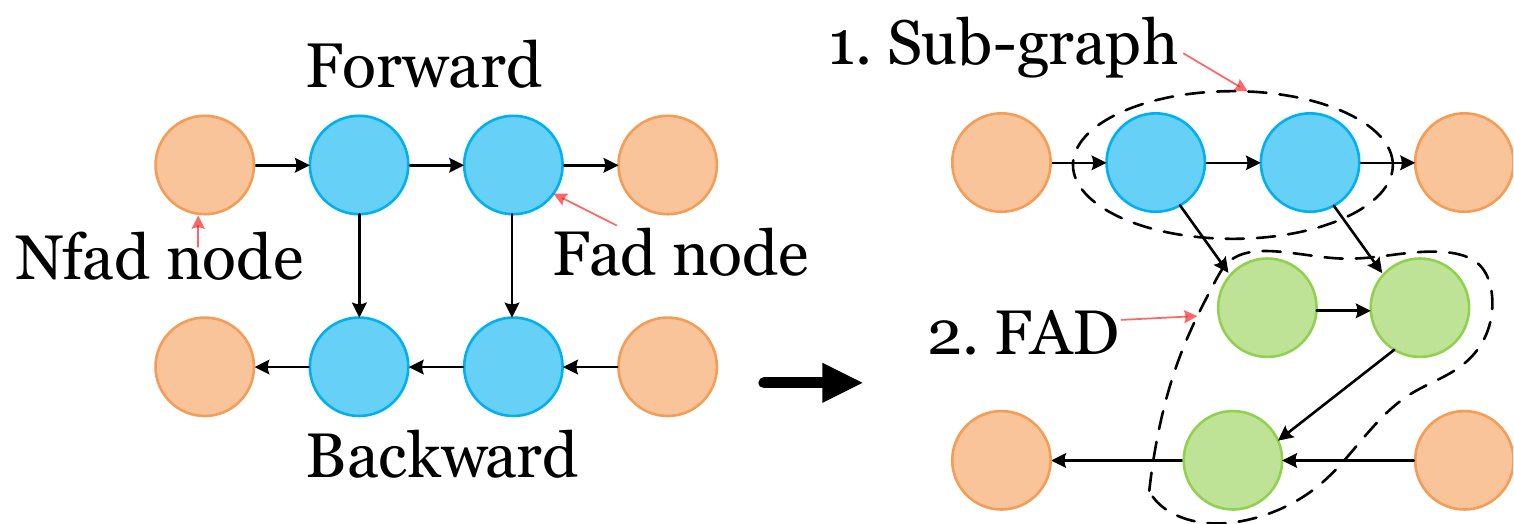}
    \caption{Static computation graph optimization.}
    \label{fig:static_graph}
\end{figure}

There are two steps with the computation graph.
First of all, TensorFlow builds a static computation graph with forward pass and backward pass. With the definitions of \benchmark{fad} operation, we can divide the forward computation graph into two types of sub-graphs: FAD and non-FAD.
Second, each sub-graph can be regarded as a node with in-degree and out-degree. We find FAD sub-graphs with 1 in-degree and $m \geq 1$ out-degree and optimize the computation. Finally, we get FAD intermediate derivatives, which will be saves for the backward gradient computation mentioned in \Fig{fig:static_graph}.

This procedure can be implemented after the gradient computation graph built by the \benchmark{backward} function before the graph automatic optimization. We eliminate the original connection between forward and backward for intermediate variables and the extra computation nodes in FAD sub-graphs, then attach FAD computation graph on the original forward computation graph and save the only single tensor of FAD derivative for the backward, finally build a new node in backward for accumulating the derivatives and computing the gradient.

After that, TensorFlow will automatically execute the computation graph optimization, such as prune nodes that do not affect the output, eliminate common subexpressions, and simplify arithmetic statements. FAD will not impact the original forward computation and reuse the tensor immediately and is friendly to accelerator memory locality.

\section{Dynamic computation graph optimizaton}
\label{sec:dynamic_graph}

The insight of FAD optimization on dynamic computation graph is similar to the static one. However, its complexity is more significant because the imperative mode can not provide enough graph information.  We design a finite state machine and an interactive approach for dynamic graph optimization to maintain the correctness and efficiency of FAD.

\subsection{Finite State Machine}
First, we need to maintain the correctness of FAD algorithm with the pattern: $\mathbb{R}^1 \rightarrow \mathbb{R}^M$. This goal is easy to reach in the static graph from the global perspective but difficult in the dynamic graph. We define T-O (tensor-operator) pairs as the basic optimization unit for the dynamic graph. The T-O pairs have two types of OPs (\benchmark{fad}/\benchmark{nfad}), and two types of input tensors: \benchmark{fad}/\benchmark{nfad} tensor is produced by a previous \benchmark{fad}/\benchmark{nfad} operator. 
Denote the \benchmark{fad} by Y and \benchmark{nfad} by N. Then, we have 4 forward states for T-O pairs: NN, NY, YY, and YN. 

\begin{figure}[t]
    \centering
    \includegraphics[width=0.8\columnwidth]{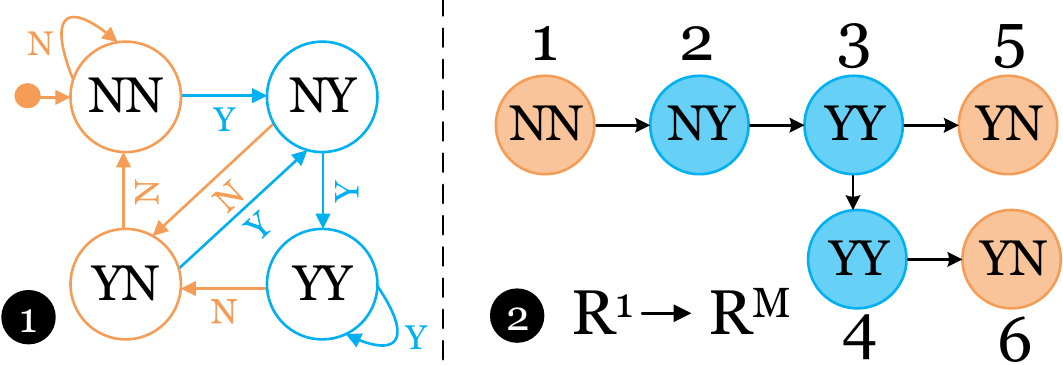}
      \vspace*{0.04cm}
    \caption{Forward pass state machine and FAD pattern.}
    \label{fig:fad_pattern}
\end{figure} 

For the imperative mode, we can regard the next OP as an input event. Therefore, we can construct a finite state machine (FSM) for forward pass depicted in \Fig{fig:fad_pattern} \circled{1}. Naturally, the DNN models can be represented by a finite forward pass state sequence, where the original BP has only NN states. According to the FSM, NY is the beginning of a FAD sequence ended by YN. Based on the \benchmark{fad} OP definition in \Sec{sec:definitions}, the NY forward mode has a single input source, and YN can be referenced by multiple OPs. Therefore, we can adopt the forward mode FSM to ensure that FAD sequence can meet the specific pattern: $\mathbb{R}^1 \rightarrow \mathbb{R}^M$.

\Fig{fig:fad_pattern} \circled{2} presents an example of a \benchmark{fad} sequence execution with $\mathbb{R}^1 \rightarrow \mathbb{R}^M$. 
The first node (Node 1) has the original NN forward mode and produce a \benchmark{nfad} output tensor transferred to the next \benchmark{fad} Node 2. Assume that Node 2 has a \benchmark{fad} operator and will produce a \benchmark{fad} output tensor. Then, we denote the Node 2 with NY forward mode, which is the beginning node of the \benchmark{fad} sequence. Then, Node 3 and Node 4 have \benchmark{fad} operators and are YY forward modes. Finally, the Node 5 and Node 6 are both end nodes with the \benchmark{nfad} and produce \benchmark{nfad} output tensors.

\begin{figure*}[t]
    \centering
    \includegraphics[width=2\columnwidth]{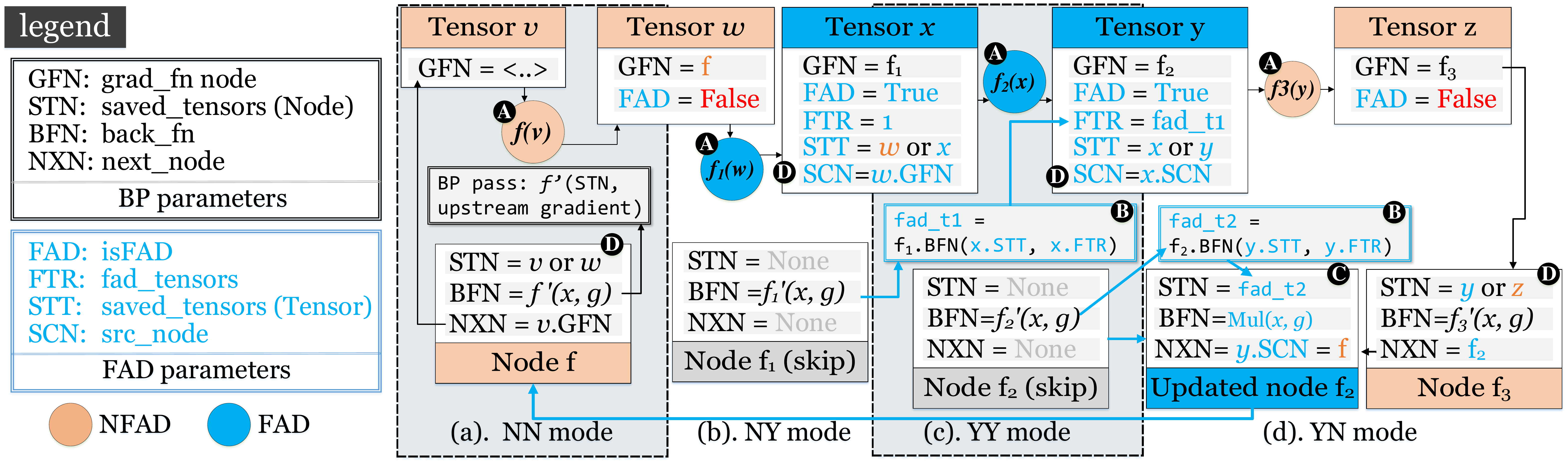}
    \vspace*{0.4cm}
    \caption{The forward pass execution. (a) Original forward mode (NN); (b) NY mode; (c)  YY mode; (d) YN mode.}
    \label{fig:forward_ad}
\end{figure*}
    
\subsection{Interactive Approach}
We should release the intermediate variables tensor appropriately and accurately to reduce the memory footprint with high memory efficiency.
For example, the imperative mode has executed the operation with tensor $x$. It is unknown whether the tensor $x$ would be used in the following program fragment. That can impact the tensor when and where to release.
Suppose we release the tensor $x$ immediately after FAD computation. That may cause serious fault when $x$ is on the critical path and referenced by the following operators. On the contrary, the tensor $x$ will impact the memory efficiency if we release the tensor too late.
Therefore, how and when to release the tensor are essential for the imperative framework.
To tackle this issue, we design an interactive approach based on the forward pass FSM with the four modes shown in \Fig{fig:forward_ad}.


\paragraph{Parameters}
There are two groups of parameters. We denote the original parameters for back-propagation in black. There are four parameters.
\begin{itemize}
    \item \textbf{{\benchmark{back\_fn}}} (BFN) is the backward function, which is implemented by the original framework;
    \item \textbf{\benchmark{saved\_tensors (node)}} (STN) is short for the saved intermediate tensors for computing the gradient using BFN;
    \item \textbf{\benchmark{grad\_node}} is a BP node with BFN and STN;
    \item \textbf{\benchmark{next\_node}} (NXN) denotes the pointer of the next BP node.
\end{itemize}

We need to add four new FAD parameters (blue) to support the forward pass FSM and interactive approach.
\begin{itemize}
    \item \textbf{\benchmark{isFAD}} (FAD) identifies the \benchmark{fad} tensor;     
    \item \textbf{\benchmark{saved\_tensors (tensor)}} (STT) is the same as STN but temporary. It can be recursively released when the tensor is released;  
    \item \textbf{\benchmark{fad\_tensors}} (FTR) is the \benchmark{fad} result tensors. Particularly, the FTR sets a scalar value $1$ for the NY (beginning) mode;      
    \item \textbf{\benchmark{src\_node}} (SCN) is the head node of a FAD sequence. The last YY nodes' NXN will point it in the post-process.
\end{itemize}

\paragraph{Execution}
There are four basic execution components and each forward mode executes the different execution components shown in \Tbl{tbl:execution} and \Fig{fig:forward_ad}. 

\circled{A} \textbf{Forward computation} execute the forward operation and is the basic unit for all modes. 

\circled{B} \textbf{Forward-AD computation} can immediately calculate the derivatives for the node with the forward differentiation mode. It is only for modes with a \benchmark{fad} tensor (YY and YN modes).
FAD computation exploits the original BFN with the STT and FTR of the input tensor, such as the \benchmark{fad\_t1} of YY mode in \Fig{fig:forward_ad}.

\circled{C} \textbf{Forward-AD post-process} is only for YN mode (end state). The post-process is only for the YN mode. It will save FAD result tensors as the STN and update BFN and NXN for the previous node. For example, the YN mode in \Fig{fig:forward_ad} computes FAD result \benchmark{fad\_t2} and saves it for the STN of Node $f_2$. Then, FAD post-process replaces the BFN with a direct \benchmark{Multiplication} with upstream gradient and updates the NXN with the head Node $f$. For a \benchmark{binary} operator, the post-process will merge (accumulate) FAD results from the two input tensors.

\circled{D} \textbf{Parameters update} is also the basic unit for all modes, which updates the node and tensor parameters. For the NN mode, it is the original update process for the BP node. The NY and YY modes only update the information with the tensor and will skip the BP nodes, which will not be used in the Forward-AD. For the YN mode, it will update the BP node based on the Forward-AD post-process.

{
\renewcommand{\arraystretch}{1.5}
\begin{table}[h]
\centering  
\resizebox{0.8\linewidth}{!}{
\begin{tabular}{c | c | c | c} 
\Xhline{1pt} 
 NN & NY & YY & YN\\\hline
 \circled{A} \circled{D} & \circled{A} \circled{D} & \circled{A} \circled{B} \circled{D} &\circled{A} \circled{B} \circled{C} \circled{D} \\
\Xhline{1pt}
\end{tabular}
}
\caption{Forward pass execution}
\label{tbl:execution}
\end{table}
}

\paragraph{Resource release} 
After 4 modes are executed sequentially, the Tensors $w$, $x$ and Node $f_1$ have no reference on them in \Fig{fig:forward_ad}. The Tensor $y$ may have a reference determind by derivative function of $f_3$.
It is known that PyTorch exploits the reference count and garbage collector mechanism of Python. PyTorch can make sure that all intermediate variables are released as soon as they become unneeded~\cite{paszke2017automatic}. When the Tensor $w$, $x$ and node $f_1$ are released, the STT and FTR referenced by Tensor $w$, $x$ are also released automatically and recursively.

\paragraph{Efficiency}
The interactive approach utilizes information from individual operators with high efficiency and robustness.
First, the approach does not influence the original forward mode (NN mode). 
Second, the approach will skip the intermediate nodes, such as Node $f_1$ and original Node $f_2$, leading to an efficient backward gradient execution. The last updated Node $f_2$ can directly pass the gradient to the head Node $f$.
Besides, if FAD sequence only has a single \benchmark{fad} operator, i.e., NY $\rightarrow$ YN, FAD post-process (\circled{C} ) in YN mode will easily skip FAD computation (\circled{B}) to avoid the extra computation with FTR $=1$.
Therefore, the approach is highly robust to the pattern $\mathbb{R}^1 \rightarrow \mathbb{R}^M, M \geq 1$ based on the FSM.

Finally, we exploit the interactive approach to optimize the dynamic computation graph for the T-O pairs and embed the nested FAD computation inside the forward pass without influence on BP. The mechanism is based on FSM and original garbage collector and maintains FAD's correctness and high memory efficiency.
\section{Evaluation}
\label{sec:evaluation}

In this section, we perform the nested Forward-AD overhead and memory efficiency evaluations in both static and dynamic computation graphs.

\subsection{Methodology}
For the implementation of the static computation graph, we modify TensorFlow and embed a hook function for FAD optimization as detailed in \Sec{sec:static_graph} before execution. Then, TensorFlow can optimize the computation graph without saving the intermediate variables.
For the implementation of the dynamic computation graph, we use the ``Autograd Function'' in PyTorch to override the specific \benchmark{fad} operators, including \benchmark{add}, \benchmark{mul}, \benchmark{exp}, \benchmark{tanh}, \benchmark{softplus}, \benchmark{sigmoid}, etc.

Owing to the space limitation, we evaluate two popular neural networks, ResNet~\cite{he2016deep} (CNN) and BERT~\cite{devlin2018bert} (Transformer), which cover tasks from the computer vision and NLP domain.
We evaluate ResNet-50 for image classification on the ImageNet~\cite{deng2009imagenet} dataset. ResNet has been widely used as the backbone in many applications, e.g., Mask R-CNN for instance segmentation~\cite{8237584}. For the state-of-the-art Transformer model family, we use BERT-base with $128$ sequence length on GLUE (general language understanding evaluation)~\cite{wang2019glue} dataset. We only present the results on four datasets (MRPC, CoLA, SST-2, and MNLI).
Finally, we implement the activation functions Mish~\cite{misra2019mish}, Swish~\cite{ramachandran2017searching}, and GELU~\cite{hendrycks2020gaussian} on the networks respectively.
All the experiments are conducted on the NVIDIA RTX 2080Ti GPU~\cite{turing2018whitepaper} with 11~GB of off-chip GDDR-based global memory. 
 
\subsection{Memory Footprint Reduction}

\begin{figure}[t]
    \centering
    \vspace*{4mm}
    \includegraphics[width=1\columnwidth]{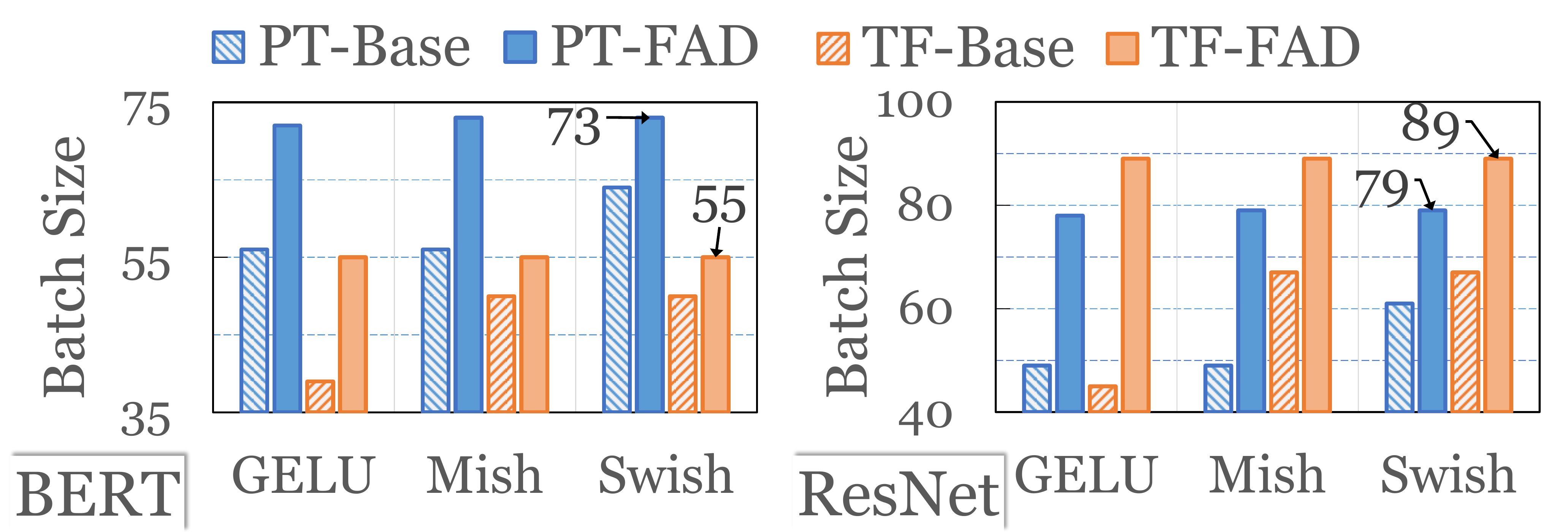}
    \caption{The batch size results with BERT-base and ResNet-50 on PyTorch (PT) and TensorFlow (TF).}
    \label{fig:batchsize}
\end{figure}

By injecting FAD into DNN frameworks, we can reduce the memory footprint and allow larger batches for the DNN training.
Therefore, we first adopt the metric of maximum supported batch size to represent the amount of memory footprint.
\Fig{fig:batchsize} depicts the maximum batch size results for the baseline and Forward-AD on TensorFlow and PyTorch. 
TensorFlow adopts the static graph optimization for the self-attention mechanism~\cite{vaswani2017attention}, which is the main component of BERT model. As such, TensorFlow has less memory usage, and TensorFlow results are better than PyTorch ones with BERT. TensorFlow and PyTorch have similar ResNet implementation, except for the activation function. Thus, ResNet evaluation has close results for both frameworks.

We observe that our proposed Forward-AD allows users to take advantage of a larger batch size with the same hardware configuration.
Compared to original frameworks, FAD promotes maximum batch size by up to \textbf{1.97$\times$} for ResNet-50. Especially for BERT, FAD also achieves up to \textbf{50\%} memory reduction than the baseline.
Finally, FAD achieves \textbf{1.34$\times$} on average than the original TensorFlow and PyTorch.
Moreover, we also implement BERT-large with 24 Transformer layers on the original GELU. Unfortunately, that occurs out-of-memory runtime error on both TensorFlow and PyTorch. 
However, Forward-AD still maintains applicability for BERT-large.

\subsection{Performance}
In this section, we evaluate the performance of Nested Forward-AD against Recomputation and original execution on Tensorflow and PyTorch.

\paragraph{Overhead}
First, we measure the runtime overhead of Forward-AD due to computation graph reconstruction.
Because FAD only changes the computation order and has no additional computation than the original model, FAD and the original model have almost the same training speed with little overhead. On the contrary, FAD can immediately exploit the output tensor of the operator to compute the temporary derivatives (gradients), which can lead to a better cache hit rate. As such, FAD can achieve light performance improvement than the original TensorFlow and PyTorch. This means the overhead introduced by Forward-AD is negligible.

\begin{figure}[t]
    \centering
    \vspace{4mm}
    \includegraphics[width=1\columnwidth]{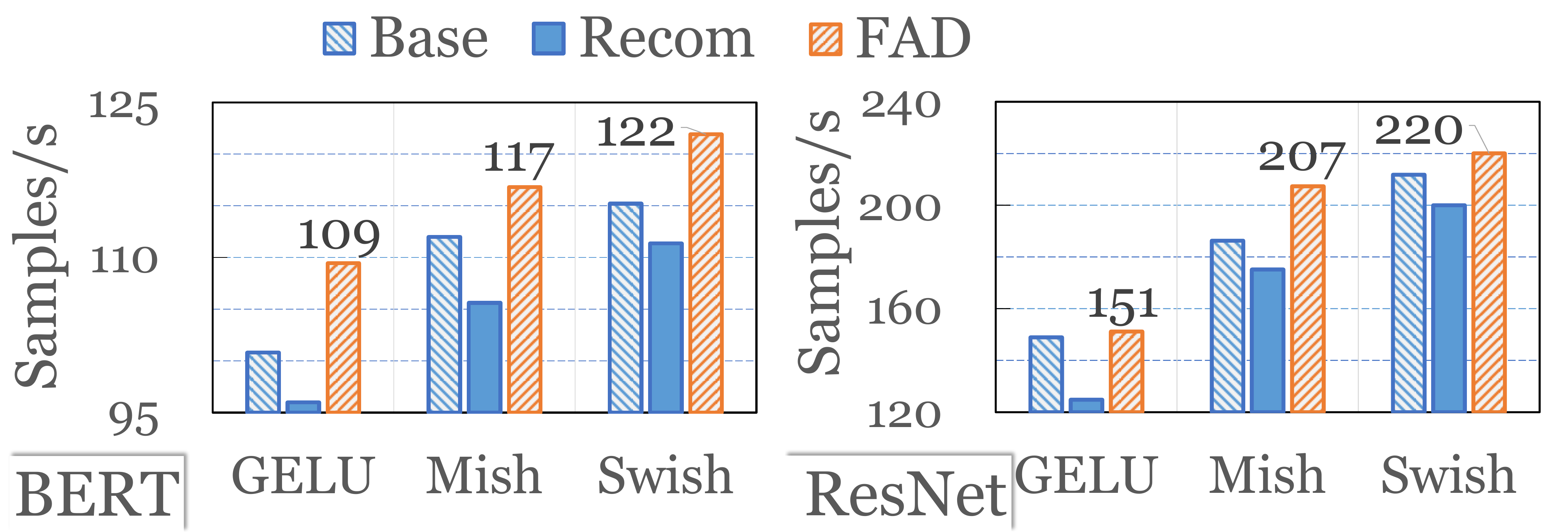}
    \caption{The end-to-end performance with training samples per second results with baseline (Base), recomputation (Recom) and forward AD (FAD).}
    \label{fig:performance}
\end{figure} 

\begin{table}[b]
    \renewcommand{\arraystretch}{1.4}
    \centering  
    \resizebox{1\linewidth}{!}{
    \begin{tabular}{c|c|c|c|c|c|c|c|c}
        \Xhline{1pt} 
         & \multicolumn{2}{c|}{MRPC} & \multicolumn{2}{c|}{CoLA} & \multicolumn{2}{c|}{SST-2} & \multicolumn{2}{c}{MNLI} \\\hline
         & Acc.      & Time     & MCC         & Time       & Acc.      & Time      & Acc.     & Time      \\\hline
    Base & 86.02         & 111s     & 59.12       & 258s       & 91.40          & 0.57h     & 83.36        & 3.34h     \\\hline
    REC  & 86.02         & 118s     & 60.06       & 277s       & 91.74         & 0.61h     & 83.54        & 3.52h     \\\hline
    FAD  & \textbf{86.02}        & \textbf{101s}     & \textbf{60.06}       & \textbf{235s}       & \textbf{91.74}         & \textbf{0.51h}     & \textbf{83.54}       & \textbf{2.95h}   \\ \hline
    \Xhline{1pt}
    \end{tabular}
    }
    \caption{The fine-tuning time and accuracy for BERT-Base model on four datasets (MRPC, CoLA, SST-2, MNLI). MCC is the Matthews correlation\cite{matthews1975comparison}.}
    \label{tbl:e2e}
\end{table}
\paragraph{Performance Comparison}
As for the performance baseline, we use the training speed under the maximum batch size.
We compare Forward-AD with recomputation (REC) and original execution on PyTorch. The end-to-end model performance (samples/second) is shown in \Fig{fig:performance}.

We find that our design achieve different performances for different activation functions. Swish has the best performance, and GELU is the worst due to GELU having the most complex element-wise computation. 
For the computation modes, recomputation leads to performance degradation due to the extra computation for regenerating the intermediate variables, as shown in \Fig{fig:fad}. The recomputation drops about 10\% performance than PyTorch baseline. 
FAD consistently demonstrates the best performance across all workloads. Finally, FAD achieves an average of 14\% and 16.5\% speedup than recomputation on BERT and ResNet, respectively.

\paragraph{End-to-end Execution Time and Accuracy}
We also evaluate the end-to-end execution time of finetuning with the pre-trained model for BERT-base with GELU on four datasets shown in \Tbl{tbl:e2e}.
In practice, the recomputation is usually used for activation function, and the complex operations, e.g., Conv and FC, will be optimized with the swapping method~\cite{peng2020capuchin}.
For these results of BERT, it is noteworthy that FAD surpasses recomputation by an average of \textbf{1.78$\times$} speedup on the execution time of the activation function, which accounts for 18.3\% of end-to-end original training time. 

Finally, FAD achieves 15\% performance improvement than recomputation on average.
We can also observe that the accuracy has a minor improvement because larger batches provide more optimization space for the DNN model.

\section{Realted work}
\label{sec:related_work}
\paragraph{Memory-efficient DNN}
Recomputation~\cite{chen2016training, jain2019checkmate, wang2018superneurons} and offloading~\cite{peng2020capuchin, rhu2016vdnn, wang2018superneurons} is the favored method to reduce the footprint in training, but with significant overhead, e.g., memory swapping and extra computation. CDMA~\cite{rhu2018compressing}, Gist~\cite{jain2018gist}, JPAC-ACT~\cite{evans2020jpeg} and buddy compression~\cite{choukse2020buddy} compress the data by leveraging the sparsity or character of the feature map in particular neural network architecture to reduce the memory burden in the training stage. For the inference stage, many quantization~\cite{nagel2020up, guo2022ant, guo2022squant} and sparsity~\cite{qin2020sigma, Qiu_2019_CVPR, gan2020ptolemy, guo2020accelerating, wang2021dual} works have been proposed to reduce the parameters and computation of DNN.

\paragraph{Automatic Differentiation}
Even though forward automatic differentiation is widely studied in mathematics~\cite{RevelsLubinPapamarkou2016} and some deep learning frameworks (such as TensorFlow~\cite{abadi2016tensorflow} and JAX~\cite{jax2018github}) have supported forward AD, they cannot dynamically nest forward AD into backward AD to update weight, users have to manually combine the separate forward/backward AD sequence.

\section{Conclusion}
\label{sec:conclude}
In this paper, we propose the nested Forward-AD in the DNN framework that reduces the memory footprint. The key insight is that FAD is much more memory efficient than BP with the specific pattern ($\mathbb{R}^1 \rightarrow \mathbb{R}^M, M \geq 1$). We deploy nested FAD in static and dynamic computation graph design with up to 1.97$\times$ memory reduction than the baseline model without the overhead.

\section*{Acknowledgment}

This work was supported by the National Key R\&D Program of China under Grant 2021ZD0110104, the National Natural Science Foundation of China (NSFC) grant (U21B2017, 62072297, and 61832006).
The authors would like to thank the anonymous reviewers for their constructive feedback for improving the work. 
Any opinions, findings, and conclusions in this paper are those of the authors only and do not necessarily reflect the views of our sponsors.


\bibliographystyle{IEEEtran}
\bibliography{main}

\begin{thebibliography}{10}
\providecommand{\url}[1]{#1}
\csname url@samestyle\endcsname
\providecommand{\newblock}{\relax}
\providecommand{\bibinfo}[2]{#2}
\providecommand{\BIBentrySTDinterwordspacing}{\spaceskip=0pt\relax}
\providecommand{\BIBentryALTinterwordstretchfactor}{4}
\providecommand{\BIBentryALTinterwordspacing}{\spaceskip=\fontdimen2\font plus
\BIBentryALTinterwordstretchfactor\fontdimen3\font minus
  \fontdimen4\font\relax}
\providecommand{\BIBforeignlanguage}[2]{{%
\expandafter\ifx\csname l@#1\endcsname\relax
\typeout{** WARNING: IEEEtran.bst: No hyphenation pattern has been}%
\typeout{** loaded for the language `#1'. Using the pattern for}%
\typeout{** the default language instead.}%
\else
\language=\csname l@#1\endcsname
\fi
#2}}
\providecommand{\BIBdecl}{\relax}
\BIBdecl

\bibitem{he2016deep}
K.~He, X.~Zhang, S.~Ren, and J.~Sun, ``Deep residual learning for image
  recognition,'' in \emph{2016 IEEE Conference on Computer Vision and Pattern
  Recognition (CVPR)}, 2016, pp. 770--778.

\bibitem{devlin2018bert}
J.~Devlin, M.-W. Chang, K.~Lee, and K.~Toutanova, ``Bert: Pre-training of deep
  bidirectional transformers for language understanding,'' \emph{arXiv preprint
  arXiv:1810.04805}, 2018.

\bibitem{vaswani2017attention}
A.~Vaswani, N.~Shazeer, N.~Parmar, J.~Uszkoreit, L.~Jones, A.~N. Gomez,
  {\L}.~Kaiser, and I.~Polosukhin, ``Attention is all you need,'' in
  \emph{Advances in neural information processing systems}, 2017, pp.
  5998--6008.

\bibitem{deng2009imagenet}
J.~Deng, W.~Dong, R.~Socher, L.-J. Li, K.~Li, and L.~Fei-Fei, ``Imagenet: A
  large-scale hierarchical image database,'' in \emph{2009 IEEE conference on
  computer vision and pattern recognition}.\hskip 1em plus 0.5em minus
  0.4em\relax Ieee, 2009, pp. 248--255.

\bibitem{manning1999foundations}
C.~Manning and H.~Schutze, \emph{Foundations of statistical natural language
  processing}.\hskip 1em plus 0.5em minus 0.4em\relax MIT press, 1999.

\bibitem{misra2019mish}
D.~Misra, ``Mish: A self regularized non-monotonic neural activation
  function,'' \emph{arXiv preprint arXiv:1908.08681}, 2019.

\bibitem{ramachandran2017searching}
P.~Ramachandran, B.~Zoph, and Q.~V. Le, ``Searching for activation functions,''
  \emph{arXiv preprint arXiv:1710.05941}, 2017.

\bibitem{hendrycks2020gaussian}
D.~Hendrycks and K.~Gimpel, ``Gaussian error linear units (gelus),''
  \emph{arXiv preprint arXiv:1606.08415}, 2016.

\bibitem{jain2019checkmate}
P.~Jain, A.~Jain, A.~Nrusimha, A.~Gholami, P.~Abbeel, K.~Keutzer, I.~Stoica,
  and J.~E. Gonzalez, ``Checkmate: Breaking the memory wall with optimal tensor
  rematerialization,'' \emph{MLSys}, 2019.

\bibitem{rhu2016vdnn}
M.~Rhu, N.~Gimelshein, J.~Clemons, A.~Zulfiqar, and S.~W. Keckler, ``vdnn:
  Virtualized deep neural networks for scalable, memory-efficient neural
  network design,'' in \emph{2016 49th MICRO}, 2016, pp. 1--13.

\bibitem{rhu2018compressing}
M.~Rhu, M.~O'Connor, N.~Chatterjee, J.~Pool, Y.~Kwon, and S.~W. Keckler,
  ``Compressing dma engine: Leveraging activation sparsity for training deep
  neural networks,'' in \emph{2018 HPCA}, 2018, pp. 78--91.

\bibitem{chen2016training}
T.~Chen, B.~Xu, C.~Zhang, and C.~Guestrin, ``Training deep nets with sublinear
  memory cost,'' \emph{arXiv preprint arXiv:1604.06174}, 2016.

\bibitem{GunesBaydin2018}
A.~G. Baydin, B.~A. Pearlmutter, A.~A. Radul, and J.~M. Siskind, ``Automatic
  differentiation in machine learning: a survey,'' \emph{Journal of Marchine
  Learning Research}, vol.~18, pp. 1--43, 2018.

\bibitem{paszke2017automatic}
A.~Paszke, S.~Gross, S.~Chintala, G.~Chanan, E.~Yang, Z.~DeVito, Z.~Lin,
  A.~Desmaison, L.~Antiga, and A.~Lerer, ``Automatic differentiation in
  pytorch,'' 2017.

\bibitem{abadi2016tensorflow}
M.~Abadi, P.~Barham, J.~Chen, Z.~Chen, A.~Davis, J.~Dean, M.~Devin,
  S.~Ghemawat, G.~Irving, M.~Isard \emph{et~al.}, ``$\{$TensorFlow$\}$: A
  system for $\{$Large-Scale$\}$ machine learning,'' in \emph{12th OSDI}, 2016,
  pp. 265--283.

\bibitem{lecun1995convolutional}
Y.~LeCun, Y.~Bengio \emph{et~al.}, ``Convolutional networks for images, speech,
  and time series,'' \emph{The handbook of brain theory and neural networks},
  vol. 3361, no.~10, p. 1995, 1995.

\bibitem{wang2019demystifying}
F.~Wang, D.~Zheng, J.~Decker, X.~Wu, G.~M. Essertel, and T.~Rompf,
  ``Demystifying differentiable programming: Shift/reset the penultimate
  backpropagator,'' \emph{Proceedings of the ACM on Programming Languages},
  vol.~3, no. ICFP, pp. 1--31, 2019.

\bibitem{naumann2008optimal}
U.~Naumann, ``Optimal jacobian accumulation is np-complete,''
  \emph{Mathematical Programming}, vol. 112, no.~2, pp. 427--441, 2008.

\bibitem{8237584}
K.~He, G.~Gkioxari, P.~Dollár, and R.~Girshick, ``Mask r-cnn,'' in \emph{2017
  IEEE International Conference on Computer Vision (ICCV)}, 2017, pp.
  2980--2988.

\bibitem{wang2019glue}
A.~{Wang}, A.~{Singh}, J.~{Michael}, F.~{Hill}, O.~{Levy}, and S.~R. {Bowman},
  ``Glue: A multi-task benchmark and analysis platform for natural language
  understanding,'' in \emph{ICLR}, 2019.

\bibitem{turing2018whitepaper}
NVIDIA, ``{NVIDIA Turing GPU Architecture Whitepaper},'' 2018.

\bibitem{matthews1975comparison}
B.~W. Matthews, ``Comparison of the predicted and observed secondary structure
  of t4 phage lysozyme,'' \emph{Biochimica et Biophysica Acta (BBA)-Protein
  Structure}, vol. 405, no.~2, pp. 442--451, 1975.

\bibitem{peng2020capuchin}
X.~Peng, X.~Shi, H.~Dai, H.~Jin, W.~Ma, Q.~Xiong, F.~Yang, and X.~Qian,
  ``Capuchin: Tensor-based gpu memory management for deep learning,'' in
  \emph{Proceedings of the 25th International Conference on Architectural
  Support for Programming Languages and Operating Systems}, 2020, pp. 891--905.

\bibitem{wang2018superneurons}
L.~Wang, J.~Ye, Y.~Zhao, W.~Wu, A.~Li, S.~L. Song, Z.~Xu, and T.~Kraska,
  ``Superneurons: Dynamic gpu memory management for training deep neural
  networks,'' in \emph{Proceedings of the 23rd ACM SIGPLAN symposium on
  principles and practice of parallel programming}, 2018, pp. 41--53.

\bibitem{jain2018gist}
A.~Jain, A.~Phanishayee, J.~Mars, L.~Tang, and G.~Pekhimenko, ``Gist: Efficient
  data encoding for deep neural network training,'' in \emph{2018 ACM/IEEE 45th
  Annual International Symposium on Computer Architecture (ISCA)}.\hskip 1em
  plus 0.5em minus 0.4em\relax IEEE, 2018, pp. 776--789.

\bibitem{evans2020jpeg}
R.~D. Evans, L.~Liu, and T.~M. Aamodt, ``Jpeg-act: accelerating deep learning
  via transform-based lossy compression,'' in \emph{2020 ACM/IEEE 47th Annual
  International Symposium on Computer Architecture (ISCA)}.\hskip 1em plus
  0.5em minus 0.4em\relax IEEE, 2020, pp. 860--873.

\bibitem{choukse2020buddy}
E.~Choukse, M.~B. Sullivan, M.~O’Connor, M.~Erez, J.~Pool, D.~Nellans, and
  S.~W. Keckler, ``Buddy compression: Enabling larger memory for deep learning
  and hpc workloads on gpus,'' in \emph{2020 ACM/IEEE 47th Annual International
  Symposium on Computer Architecture (ISCA)}.\hskip 1em plus 0.5em minus
  0.4em\relax IEEE, 2020, pp. 926--939.

\bibitem{nagel2020up}
M.~Nagel, R.~A. Amjad, M.~Van~Baalen, C.~Louizos, and T.~Blankevoort, ``Up or
  down? adaptive rounding for post-training quantization,'' in
  \emph{International Conference on Machine Learning}.\hskip 1em plus 0.5em
  minus 0.4em\relax PMLR, 2020, pp. 7197--7206.

\bibitem{guo2022ant}
C.~Guo, C.~Zhang, J.~Leng, Z.~Liu, F.~Yang, Y.~Liu, M.~Guo, and Y.~Zhu, ``Ant:
  Exploiting adaptive numerical data type for low-bit deep neural network
  quantization,'' \emph{arXiv preprint arXiv:2208.14286}, 2022.

\bibitem{guo2022squant}
\BIBentryALTinterwordspacing
C.~Guo, Y.~Qiu, J.~Leng, X.~Gao, C.~Zhang, Y.~Liu, F.~Yang, Y.~Zhu, and M.~Guo,
  ``{SQ}uant: On-the-fly data-free quantization via diagonal hessian
  approximation,'' in \emph{International Conference on Learning
  Representations}, 2022. [Online]. Available:
  \url{https://openreview.net/forum?id=JXhROKNZzOc}
\BIBentrySTDinterwordspacing

\bibitem{qin2020sigma}
E.~Qin, A.~Samajdar, H.~Kwon, V.~Nadella, S.~Srinivasan, D.~Das, B.~Kaul, and
  T.~Krishna, ``Sigma: A sparse and irregular gemm accelerator with flexible
  interconnects for dnn training,'' in \emph{2020 IEEE International Symposium
  on High Performance Computer Architecture (HPCA)}.\hskip 1em plus 0.5em minus
  0.4em\relax IEEE, 2020, pp. 58--70.

\bibitem{Qiu_2019_CVPR}
Y.~Qiu, J.~Leng, C.~Guo, Q.~Chen, C.~Li, M.~Guo, and Y.~Zhu, ``Adversarial
  defense through network profiling based path extraction,'' in
  \emph{Proceedings of the IEEE/CVF Conference on Computer Vision and Pattern
  Recognition (CVPR)}, June 2019.

\bibitem{gan2020ptolemy}
Y.~Gan, Y.~Qiu, J.~Leng, M.~Guo, and Y.~Zhu, ``Ptolemy: Architecture support
  for robust deep learning,'' in \emph{2020 53rd Annual IEEE/ACM International
  Symposium on Microarchitecture (MICRO)}.\hskip 1em plus 0.5em minus
  0.4em\relax IEEE, 2020, pp. 241--255.

\bibitem{guo2020accelerating}
C.~Guo, B.~Y. Hsueh, J.~Leng, Y.~Qiu, Y.~Guan, Z.~Wang, X.~Jia, X.~Li, M.~Guo,
  and Y.~Zhu, ``Accelerating sparse dnn models without hardware-support via
  tile-wise sparsity,'' in \emph{SC20: International Conference for High
  Performance Computing, Networking, Storage and Analysis}.\hskip 1em plus
  0.5em minus 0.4em\relax IEEE, 2020, pp. 1--15.

\bibitem{wang2021dual}
Y.~Wang, C.~Zhang, Z.~Xie, C.~Guo, Y.~Liu, and J.~Leng, ``Dual-side sparse
  tensor core,'' in \emph{2021 ACM/IEEE 48th Annual International Symposium on
  Computer Architecture (ISCA)}.\hskip 1em plus 0.5em minus 0.4em\relax IEEE,
  2021, pp. 1083--1095.

\bibitem{RevelsLubinPapamarkou2016}
J.~{Revels}, M.~{Lubin}, and T.~{Papamarkou}, ``Forward-mode automatic
  differentiation in {J}ulia,'' \emph{arXiv:1607.07892 [cs.MS]}, 2016.

\bibitem{jax2018github}
\BIBentryALTinterwordspacing
J.~Bradbury \emph{et~al.}, ``{JAX}: composable transformations of
  {P}ython+{N}um{P}y programs,'' 2018. [Online]. Available:
  \url{http://github.com/google/jax}
\BIBentrySTDinterwordspacing

\end{thebibliography}

\end{document}